\documentstyle[aclap]{article}
\def\ie{{\em i.e.}}
\def\eg{{\em e.g.}}

\def\A{$\forall$}
\def\E{$\exists$}

\def\N{$\neg$}
\def\AND{$\wedge$}

\def\HAT{$^{\wedge}$}
\def\s{$\ast$}
\def\U{\"{u}}

\title{\vspace{-0.5in}Morphological Cues for Lexical Semantics} 

\author{Marc Light \\
Seminar f\"{u}r Sprachwissenschaft \\
Universit\"{a}t T\"{u}bingen \\ 
Wilhelmstr. 113 \\ 
D-72074 T\"{u}bingen \\ Germany \\
{\tt light@sfs.nphil.uni-tuebingen.de}
}

\begin{document}
\maketitle
\vspace{-0.5in}
\begin{abstract}
  Most natural language processing tasks require lexical semantic
  information.  Automated acquisition of this information would thus
  increase the robustness and portability of NLP systems.  This paper
  describes an acquisition method which makes use of fixed
  correspondences between derivational affixes and lexical semantic
  information.  One advantage of this method, and of other methods
  that rely only on surface characteristics of language, is that the
  necessary input is currently available.
\end{abstract}

\section{Introduction}

Some natural language processing (NLP) tasks can be performed with
only coarse-grained semantic information about individual words.  For
example, a system could utilize word frequency and a word cooccurrence
matrix in order to perform information retrieval.  However, many NLP
tasks require at least a partial understanding of every sentence or
utterance in the input and thus have a much greater need for lexical
semantics.  Natural language generation, providing a natural language
front end to a database, information extraction, machine translation,
and task-oriented dialogue understanding all require lexical
semantics.  The lexical semantic information commonly utilized
includes verbal argument structure and selectional restrictions,
corresponding nominal semantic class, verbal aspectual class, synonym
and antonym relationships between words, and various verbal semantic
features such as causation and manner.

Machine readable dictionaries do not include much of this information
and it is difficult and time consuming to encode it by hand.  As a
consequence, current NLP systems have only small lexicons and thus can
only operate in restricted domains.  Automated methods for acquiring
lexical semantics could increase both the robustness and the
portability of such systems.  In addition, such methods might provide
insight into human language acquisition.

After considering different possible approaches to acquiring lexical
semantic information, this paper concludes that a ``surface cueing''
approach is currently the most promising.  It then introduces
morphological cueing, a type of surface cueing, and discusses an
implementation.  It concludes by evaluating morphological cues with
respect to a list of desiderata for good surface cues.

\section{Approaches to Acquiring Lexical Semantics}

One intuitively appealing idea is that humans acquire the meanings of
words by relating them to semantic representations resulting from
perceptual or cognitive processing.  For example, in a situation where
the father says {\em Kim is throwing the ball\/} and points at Kim who
is throwing the ball, a child might be able learn what {\em throw\/}
and {\em ball\/} mean.  In the human language acquisition literature,
Grimshaw (1981) and Pinker (1989) advocate this approach; others have
described partial computer implementations: Pustejovsky (1988) and
Siskind (1990).  However, this approach cannot yet provide for the
automatic acquisition of lexical semantics for use in NLP systems,
because the input required must be hand coded: no current artificial
intelligence system has the perceptual and cognitive capabilities
required to produce the needed semantic representations.

Another approach would be to use the semantics of surrounding words in
an utterance to constrain the meaning of an unknown word.  Borrowing
an example from Pinker (1994), upon hearing {\em I glipped the paper
  to shreds}, one could guess that the meaning of {\em glib\/} has
something to do with tearing.  Similarly, one could guess that {\em
  filp\/} means something like {\em eat\/} upon hearing {\em I filped
  the delicious sandwich and now I'm full}.  These guesses are cued by
the meanings of {\em paper, shreds, sandwich, delicious, full,\/} and
the partial syntactic analysis of the utterances that contain them.
Granger (1977), Berwick (1983), and Hastings (1994) describe
computational systems that implement this approach.  However, this
approach is hindered by the need for a large amount of initial lexical
semantic information and the need for a robust natural language
understanding system that produces semantic representations as output,
since producing this output requires precisely the lexical semantic
information the system is trying to acquire.

A third approach does not require any semantic information related to
perceptual input or the input utterance.  Instead it makes use of
fixed correspondences between surface characteristics of language
input and lexical semantic information: surface characteristics serve
as cues for lexical semantics of the words.  For example, if a verb is
seen with a noun phrase subject and a sentential complement, it often
has verbal semantics involving spatial perception and cognition, \eg ,
{\em believe, think, worry}, and {\em see} \cite{fisher91,gleitman90}.
Similarly, the occurrence of a verb in the progressive tense can be
used as a cue for the non-stativeness of the verb \cite{dorr92};
stative verbs cannot appear in the progress tense (\eg ,*{\em Mary is
  loving her new shoes\/}).  Another example is the use of patterns
such as $NP {, NP}* {,} and\; other NP$ to find lexical semantic
information such as hyponym \cite{hearst92}.  {\em Temples,
  treasuries, and other important civic buildings} is an example of
this pattern and from it the information that temples and treasuries
are types of civic buildings would be cued.  Finally, inducing lexical
semantics from distributional data (\eg , \cite{brown92,church_89b})
is also a form of surface cueing.  It should be noted that the set of
fixed correspondences between surface characteristics and lexical
semantic information, at this point, have to be acquired through the
analysis of the researcher---the issue of how the fixed
correspondences can be automatically acquired will not be addressed
here.

The main advantage of the surface cueing approach is that the input
required is currently available: there is an ever increasing supply of
online text, which can be automatically part-of-speech tagged,
assigned shallow syntactic structure by robust partial parsing
systems, and morphologically analyzed, all without any prior lexical
semantics.  

A possible disadvantage of surface cueing is that surface cues for a
particular piece of lexical semantics might be difficult to uncover or
they might not exist at all.  In addition, the cues might not be
present for the words of interest.  Thus, it is an empirical question
whether easily identifiable abundant surface cues exist for the needed
lexical semantic information.  The next section explores the
possibility of using derivational affixes as surface cues for lexical
semantics.

\section{Morphological Cues for Lexical Semantic Information}

Many derivational affixes only apply to bases with certain semantic
characteristics and only produce derived forms with certain semantic
characteristics.  For example, the verbal prefix {\em un-} applies to
telic verbs and produces telic derived forms.  Thus, it is possible to
use {\em un-\/} as a cue for telicity.  By searching a sufficiently
large corpus we should be able to identify a number of telic verbs.
Examples from the Brown corpus include {\em clasp, coil, fasten,
  lace}, and {\em screw}.

A more implementation-oriented description of the process is the
following: (i) analyze affixes by hand to gain fixed correspondences
between affix and lexical semantic information (ii) collect a large
corpus of text, (iii) tag it with part-of-speech tags, (iv)
morphologically analyze its words, (v) assign word senses to the base
and the derived forms of these analyses, and (vi) use this
morphological structure plus fixed correspondences to assign semantics
to both the base senses and the derived form senses.  Step (i) amounts
to doing a semantic analysis of a number of affixes the goal of which
is to find semantic generalizations for an affix that hold for a large
percentage of its instances.  Finding the right generalizations and
stating them explicitly can be time consuming but is only performed
once.  Tagging the corpus is necessary to make word sense
disambiguation and morphological analysis easier.  Word sense
disambiguation is necessary because one needs to know which sense of
the base is involved in a particular derived form, more specifically,
to which sense should one assign the feature cued by the affix.  For
example, {\em stress\/} can be either a noun {\em the stress on the
  third syllable\/} or a verb {\em the advisor stressed the importance
  of finishing quickly}.  Since the suffix {\em -ful\/} applies to
nominal bases, only a noun reading is possible as the stem of {\em
  stressful} and thus one would attach the lexical semantics cued by
{\em -ful\/} to the noun sense.  However, {\em stress\/} has multiple
readings even as a noun: it also has the reading exemplified by {\em
  the new parent was under a lot of stress}.  Only this reading is
possible for {\em stressful}.

In order to produce the results presented in the next section, the
above steps were performed as follows.  A set of 18 affixes were
analyzed by hand providing the fixed correspondences between cue and
semantics.  The cued lexical semantic information was axiomatized
using Episodic Logic \cite{hwang93}, a situation-based extension of
standard first order logic.  The Penn Treebank version of the Brown
corpus \cite{marcus_93a} served as the corpus.  Only its words and
part-of-speech tags were utilized.  Although these tags were corrected
by hand, part-of-speech tagging can be automatically performed with an
error rate of 3 to 4 percent \cite{merialdo_94,brill_94}.  The Alvey
morphological analyzer \cite{ritchie92} was used to assign
morphological structure.  It uses a lexicon with just over 62,000
entries.  This lexicon was derived from a machine readable dictionary
but contains no semantic information.  Word sense disambiguation for
the bases and derived forms that could not be resolved using
part-of-speech tags was not performed.  However, there exist systems
for such word sense disambiguation which do not require explicit
lexical semantic information \cite{yarowsky_93,schuetze_92}.

Let us consider an example.  One sense of the suffix {\em -ize\/}
applies to adjectival bases (\eg , {\em centralize\/}).  This sense of
the affix will be referred to as {\em -Aize}.  (A related but
different sense applies to nouns, \eg , {\em glamorize}.  The
part-of-speech of the base is used to disambiguate these two senses of
{\em -ize}.)  First, the regular expressions ``{\tt
  .*IZ(E$|$ING$|$ES$|$ED)\$}'' and ``{\tt \HAT V.*}'' are used to
collect tokens from the corpus that were likely to have been derived
using {\em -ize}.  The Alvey morphological analyzer is then applied to
each type.  It strips off {\em -Aize\/} from a word if it can find an
entry with a reference form of the appropriate orthographic shape and
has the features ``uninflected,'' ``latinate,'' and ``adjective.''  It
may also build an appropriate base using other affixes, \eg ,[[{\em
  tradition} {\em -al}] {\em -Aize}].\footnote{In an alternative
  version of the method, the morphological analyzer is also able to
  construct a base on its own when it is unable to find an appropriate
  base in its lexicon.  However, these ``new'' bases seldom correspond
  to actual words and thus the results presented here were derived
  using a morphological analyzer configured to only use bases that are
  directly in its lexicon or can be constructed from words in its
  lexicon.} Finally, all derived forms are assigned the lexical
semantic feature {\sc change-of-state} and all the bases are assigned
the lexical semantic feature {\sc ize-dependent}.  Only the {\sc
  change-of-state} feature will be discussed here.  It is defined by
the axiom below.  {\tt
  \begin{tabbing}
    For all predicates P with features \\
    {\sc change-of-state} and {\sc dyadic}:\\
    \A x,y,e [\=P(x,y)\s\s e -> \\
    \>        [\=\E e1:\=[\=at-end-of(e1,e) \AND \\
    \>\>\>\>              cause(e,e1)]\\
    \>\>                       [rstate(P)(y)\s\s e1] \AND \\
    \>\>\E e2:at-beginning-of(e2,e)  \\
    \>\> [\N rstate(P)(y)\s\s e2]]]
\end{tabbing}}

The operator {\tt **} is analogous to $\models$ in situation
semantics; it indicates, among other things, that a formula describes
an event.  {\tt P} is a place holder for the semantic predicate
corresponding to the word sense which has the feature. It is assumed
that each word sense corresponds to a single semantic predicate.  The
axiom states that if a {\sc change-of-state} predicate describes an
event, then the result state of this predicate holds at the end of
this event and that it did not hold at the beginning, \eg , if one
wants to formalize something it must be non-formal to begin with and
will be formal after.  The result state of an {\em -Aize\/} predicate
is the predicate corresponding to its base; this is stated in another
axiom.

Precision figures for the method were collected as follows.  The
method returns a set of normalized (\ie , uninflected) word/feature
pairs.  A human then determines which pairs are ``correct'' where
correct means that the axiom defining the feature holds for the
instances (tokens) of the word (type).  Because of the lack of word
senses, the semantics assigned to a particular word is only considered
correct, if it holds for all senses occurring in the relevant derived
word tokens.\footnote{Although this definition is required for many
  cases, in the vast majority of the cases, the derived form and its
  base have only one possible sense (\eg , {\em stressful}).} For
example, the axiom above must hold for all senses of {\em centralize}
occurring in the corpus in order for the {\em centralize}/{\sc
  change-of-state} pair to be correct.  The axiom for {\sc
  ize-dependent} must hold only for those senses of {\em central\/}
that occur in the tokens of {\em centralize\/} for the {\em
  central\/}/{\sc ize-dependent} pair to be correct.  This definition
of correct was constructed, in part, to make relatively quick human
judgements possible.  It should also be noted that the semantic
judgements require that the semantics be expressed in a precise way.
This discipline is enforced in part by requiring that the features be
axiomatized in a denotational logic.  Another argument for such an
axiomatization is that many NLP systems utilize a denotational logic
for representing semantic information and thus the axioms provide a
straightforward interface to the lexicon.

To return to our example, as shown in Table~1, there were 63 {\em
  -Aize\/} derived words (types) of which 78 percent conform to the
{\sc change-of-state} axiom.  Of the bases, 80 percent conform to the
{\sc ize-dependent} axiom which will be discussed in the next section.
Among the conforming words were {\em equalize, stabilize, {\em and}
  federalize}.  Two words that seem to be derived using the {\em
  -ize\/} suffix but do not conform to the {\sc change-of-state} axiom
are {\em penalize {\em and} socialize (with the guests)}.  A different
sort of non-conformity is produced when the morphological analyzer
finds a spurious parse.  For example, it analyzed {\em subsidize} as
{\em [sub- [side -ize]]} and thus produced the {\em sidize}/{\sc
  change-of-state} pair which for the relevant tokens was incorrect.
In the first sort, the non-conformity arises because the cue does not
always correspond to the relevant lexical semantic information.  In
the second sort, the non-conformity arises because a cue has been
found where one does not exist.  A system that utilizes a lexicon so
constructed is interested primarily in the overall precision of the
information contained within and thus the results presented in the
next section conflate these two types of false positives.

\section{Results}

This section starts by discussing the semantics of 18 derivational
affixes: {\em re-, un-, de-, -ize, -en, -ify, -le,
  -ate, -ee, -er, -ant, -age, -ment, mis-, -able, -ful, -less}, and
{\em -ness}.  Following this discussion, a table of precision statistics
for the performance of these surface cues is presented.  Due to space
limitations, the lexical semantics cued by these affixes can only be
loosely specified.  However, they have been axiomatized in a fashion
exemplified by the {\sc change-of-state} axiom above (see
\cite{light96a,light92}). 

The verbal prefixes {\em un-, de-}, and {\em re-\/} cue aspectual
information for their base and derived forms.  Some examples from the
Brown corpus are {\em unfasten, unwind, decompose, defocus,
  reactivate}, and {\em readapt}.  Above it was noted that {\em un-\/}
is a cue for telicity.  In fact, both {\em un-\/} and {\em de-\/} cue
the {\sc change-of-state} feature for their base and derived
forms---the {\sc change-of-state} feature entails the {\sc telic}
feature.  In addition, for {\em un-\/} and {\em de-}, the result state
of the derived form is the negation of the result state of the base
({\sc neg-of-base-is-rstate}), \eg , the result of unfastening
something is the opposite of the result of fastening it.  As shown by
examples like {\em reswim the last lap}, {\em re-\/} only cues the
{\sc telic} feature for its base and derived forms: the lap might have
been swum previously and thus the negation of the result state does
not have to have held previously \cite{dowty79}.  For {\em re-}, the
result state of the derived form is the same as that of the base ({\sc
  rstate-eq-base-rstate}), \eg , the result of reactivating something
is the same as activating it.  In fact, if one reactivates something
then it is also being activated: the derived form entails the base
({\sc entails-base}).  Finally, for {\em re-}, the derived form
entails that its result state held previously, \eg , if one
recentralizes something then it must have been central at some point
previous to the event of recentralization ({\sc presups-rstate}).

The suffixes {\em -Aize, -Nize, -en, -Aify, -Nify\/} all cue the {\sc
  change-of-state} feature for their derived form as was discussed for
{\em -Aize\/} above.  Some exemplars are {\em centralize, formalize,
  categorize, colonize, brighten, stiffen, falsify, intensify,
  mummify}, and {\em glorify}.  For {\em -Aize, -en\/} and {\em
  -Aify\/} a bit more can be said about the result state: it is the
base predicate ({\sc rstate-eq-base}), \eg , the result of formalizing
something is that it is formal.  Finally {\em -Aize, -en}, and {\em
  -Aify\/} cue the following feature for their bases: if a state holds
of some individual then either an event described by the derived form
predicate occurred previously or the predicate was always true of the
individual ({\sc ize-dependent}), \eg , if something is central then
either it was centralized or it was always central.

The ``suffixes'' {\em -le\/} and {\em -ate\/} should really be called
verbal endings since they are not suffixes in English, \ie , if one
strips them off one is seldom left with a word.  (Consequently, only
regular expressions were used to collect types; the morphological
analyzer was not used.)  Nonetheless, they cue lexical semantics and
are easily identified.  Some examples are {\em chuckle, dangle,
  alleviate}, and {\em assimilate}. The ending {\em -ate\/} cues a
{\sc change-of-state} verb and {\em -le\/} an {\sc activity} verb.

The derived forms produced by {\em -ee, -er}, and {\em -ant\/} all
refer to participants of an event described by their base ({\sc part-in-e}).
Some examples are {\em appointee, deportee, blower, campaigner,
  assailant}, and {\em claimant}.  In addition, the derived form of
{\em -ee\/} is also sentient of this event and non-volitional with
respect to it \cite{barker95}.  

The nominalizing suffixes {\em -age\/} and {\em -ment\/} both
produce derived forms that refer to something resulting from an event
of the verbal base predicate.  Some examples are {\em blockage,
  seepage, marriage, payment, restatement, shipment}, and {\em
  treatment}.  The derived forms of {\em -age\/} entail that an event
occurred and refer to something resulting from it 
({\sc event-and-resultant})), \eg ,
seepage entails that seeping took place and that the seepage resulted
from this seeping.  Similarly, the derived forms of {\em -ment\/}
entail that an event took place and refer either to this event, the
proposition that the event occurred, or something resulting from the
event ({\sc refers-to-e-or-prop-or-result}), \eg , a restatement entails that a restating
occurred and refers either to this event, the proposition that the
event occurred, or to the actual utterance or written document
resulting from the restating event.  (This analysis is based on 
\cite{zucchi_89}.) 

The verbal prefix {\em mis-}, \eg , {\em miscalculate\/} and {\em
  misquote}, cues the feature that an action is performed in an
incorrect manner ({\sc incorrect-manner}).  The suffix {\em -able\/}
cues a feature that it is possible to perform some action ({\sc
  able-to-be-performed}), \eg , something is enforceable if it is
possible that something can enforce it \cite{dowty79}.  The words
derived using {\em -ness\/} refer to a state of something having the
property of the base ({\sc state-of-having-prop-of-base}), \eg , in
{\em Kim's fierceness at the meeting yesterday was unusual\/} the word
{\em fierceness\/} refers to a state of Kim being fierce.  The suffix
{\em -ful\/} marks its base as abstract ({\sc abstract}): {\em
  careful, peaceful, powerful}, etc.  In addition, it marks its
derived form as the antonym of a form derived by {\em -less\/} if it
exists ({\sc less-antonym}).  The suffix {\em -less\/} marks its
derived forms with the analogous feature ({\sc ful-antonym}).  Some
examples are {\em colorful/less, fearful/less, harmful/less}, and {\em
  tasteful/less}.

The precision statistics for the individual lexical semantic features
discussed above are presented in Table~1 and Table~2.  Lexical
semantic information was collected for 2535 words (bases and derived
forms).  One way to summarize these tables is to calculate a single
precision number for all the features in a table, \ie , average the
number of correct types for each affix, sum these averages, and then
divide this sum by the total number of types.  Using this statistic it
can be said that if a random word is derived, its features have a 76
percent chance of being true and if it is a stem of a derived form,
its features have a 82 percent chance of being true.

Computing recall requires finding all true tokens of a cue.  This is a
labor intensive task.  It was performed for the verbal prefix {\em
  re-\/} and the recall was found to be 85 percent.  The majority of
the missed {\em re-\/} verbs were due to the fact that the system only
looked at verbs starting with {\tt RE} and not other
parts-of-speech, \eg , many nominalizations such as {\em
  reaccommodation\/} contain the {\em re-} morphological cue.
However, increasing recall by looking at all open class categories
would probably decrease precision.  Another cause of reduced recall is
that some stems were not in the Alvey lexicon or could not be properly
extracted by the morphological analyzer.  For example, {\em -Nize}
could not be stripped from {\em hypothesize\/} because Alvey failed to
reconstruct {\em hypothesis\/} from {\em hypothes}.  However, for the
affixes discussed here, 89 percent of the bases were present in the
Alvey lexicon.

\begin{table}[htbp]
\centering
{\footnotesize
\begin{tabular}{|l|r|r|r|}
\hline
\multicolumn{1}{|c|}{Feature} &
\multicolumn{1}{|c|}{Affix} &
\multicolumn{1}{|c|}{Types} &
\multicolumn{1}{|c|}{Precision} \\ \hline \hline
{\sc telic} & {\em re-\/} & 164 & 91\%           \\ \hline 
{\sc rstate-eq-base-} & {\em re-\/} & 164 & 65\%     \\ 
{\sc rstate} &&&\\\hline
{\sc entails-base} & {\em re-\/} & 164 & 65\%     \\ \hline 
{\sc presups-rstate} & {\em re-\/} & 164 & 65\%        \\ \hline 
{\sc change-of-state} & {\em un-\/} & 23 & 100\%             \\ \hline
{\sc neg-of-base-is-}& {\em un-\/} & 23 & 91\%  \\ 
{\sc rstate} &&& \\\hline
{\sc change-of-state} & {\em de-\/} & 35 & 34\%              \\ \hline 
{\sc neg-of-base-is-} & {\em de-\/} & 35 & 20\%  \\ 
{\sc rstate} &&& \\\hline
{\sc change-of-state} & {\em -Aize\/} & 63 & 78\%            \\ \hline 
{\sc rstate-eq-base} & {\em -Aize\/} & 63 & 75\% \\ \hline 
{\sc change-of-state} & {\em -Nize\/} & 86 & 56\%            \\ \hline
{\sc activity} & {\em -le\/} & 71 & 55\%         \\ \hline
{\sc change-of-state} & {\em -en\/} & 36 & 100\%             \\ \hline 
{\sc rstate-eq-base} & {\em -en\/} & 36 & 97\%   \\ \hline 
{\sc change-of-state} & {\em -Aify\/} & 17 & 94\%            \\ \hline
{\sc rstate-eq-base} & {\em -Aify\/} & 17 & 58\% \\ \hline
{\sc change-of-state} & {\em -Nify\/} & 21 & 67\%            \\ \hline  
{\sc change-of-state} & {\em -ate\/} & 365 & 48\%            \\ \hline 
{\sc part-in-e} & {\em -ee\/} & 22 & 91\%        \\ \hline
{\sc sentient} & {\em -ee\/} & 22 & 82\%         \\ \hline 
{\sc non-volitional} & {\em -ee\/} & 22 & 68\%          \\ \hline
{\sc part-in-e} & {\em -er\/} & 471 & 85\%       \\ \hline 
{\sc part-in-e} & {\em -ant\/} & 21 & 81\%       \\ \hline 
{\sc event-and-} & {\em -age\/} & 43 & 58\%         \\ 
{\sc resultant} &&&\\\hline
{\sc refers-to-e-or-}& {\em -ment\/} & 166 & 88\%   \\ 
{\sc prop-or-resultant}  &&& \\\hline
{\sc incorrect-manner} & {\em mis-\/} & 21 & 86\%           \\ \hline
{\sc able-to-be-} & {\em -able\/} & 148 & 84\%          \\ 
{\sc performed} &&&\\\hline
{\sc state-of-having-} & {\em -ness\/} & 307 & 97\%          \\ 
{\sc prop-of-base} &&&\\\hline
{\sc ful-antonym} & {\em -less\/} & 22 & 77\%         \\ \hline 
{\sc less-antonym} & {\em -ful\/} & 22 & 77\%         \\ \hline
\end{tabular}}
  \label{statsderived}
\caption{Derived words}
\end{table}

\begin{table}[htbp]
\centering
{\footnotesize
\begin{tabular}{|r|r|r|r|}
\hline
\multicolumn{1}{|c|}{Feature} &
\multicolumn{1}{|c|}{Affix} &
\multicolumn{1}{|c|}{Types} &
\multicolumn{1}{|c|}{Precision}   \\ \hline \hline
{\sc telic} & {\em re-\/} & 164 & 91\%   \\ \hline 
{\sc change-of-state} & {\em Vun-\/} & 23 & 91\%     \\ \hline
{\sc change-of-state} & {\em Vde-\/} & 33 & 36\%     \\ \hline
{\sc ize-dependent} & {\em -Aize\/} & 64 & 80\%\\ \hline
{\sc ize-dependent} & {\em -en\/} & 36 & 72\%\\ \hline
{\sc ize-dependent} & {\em -Aify\/} & 15 & 40\%\\ \hline
{\sc abstract} & {\em -ful\/} & 76 & 93\%    \\ \hline 
\end{tabular}}
  \label{statsbase}
\caption{Base words}
\end{table}

\section{Evaluation}

Good surface cues are easy to identify, abundant, and correspond to the
needed lexical semantic information (Hearst (1992) identifies
a similar set of desiderata).  With respect to these desiderata,
derivational morphology is both a good cue and a bad cue.

Let us start with why it is a bad cue: there may be no derivational
cues for the lexical semantics of a particular word.  This is not the
case for other surface cues, \eg , distributional cues exist for every
word in a corpus.  In addition, even if a derivational cue does exist,
the reliability (on average approximately 76 percent) of the lexical
semantic information is too low for many NLP tasks.  This
unreliability is due in part to the inherent exceptionality of lexical
generalization and thus can be improved only partially.

However, derivational morphology is a good cue in the following ways.
It provides exactly the type of lexical semantics needed for many NLP
tasks: the affixes discussed in the previous section cued nominal
semantic class, verbal aspectual class, antonym relationships between
words, sentience, etc.  In addition, working with the Brown corpus
(1.1 million words) and 18 affixes provided such information for over
2500 words.  Since corpora with over 40 million words are common and
English has over 40 common derivational affixes, one would expect to
be able to increase this number by an order of magnitude.  In
addition, most English words are either derived themselves or serve as
bases of at least one derivational affix.\footnote{The following
  experiment supports this claim.  Just over 400 open class words were
  picked randomly from the Brown corpus and the derived forms were
  marked by hand.  Based on this data, a random open class word in the
  Brown corpus has a 17 percent chance of being derived, a 56 percent
  chance of being a stem of a derived form, and an 8 percent chance of
  being both.} Finally, for some NLP tasks, 76 percent reliability may
be adequate.  In addition, some affixes are much more reliable cues
than others and thus if higher reliability is required then only the
affixes with high precision might be used.

The above discussion makes it clear that morphological cueing provides
only a partial solution to the problem of acquiring lexical semantic
information.  However, as mentioned in section 2 there are many types
of surface cues which correspond to a variety of lexical semantic
information.  A combination of cues should produce better precision
where the same information is indicated by multiple cues.  For
example, the morphological cue {\em re-\/} indicates telicity and as
mentioned above, the syntactic cue the progressive tense indicates
non-stativity \cite{dorr92}.  Since telicity is a type of
non-stativity, the information is mutually supportive.  In addition,
using many different types of cues should provide a greater variety of
information in general.  Thus morphological cueing is best seen as one
type of surface cueing that can be used in combination with others to
provide lexical semantic information.

\section{Acknowledgements}

A portion of this work was performed at the University of Rochester
Computer Science Department and supported by ONR/ARPA research grant
number N00014-92-J-1512.

\bibliographystyle{fullname}

\end{document}